\documentclass[runningheads]{llncs}

\usepackage[most]{tcolorbox}
\usepackage{amsfonts}
\usepackage{booktabs}
\usepackage{pifont}
\usepackage{colortbl}
\usepackage{graphicx}
\usepackage{amsmath}
\usepackage{amsfonts}
\usepackage{amssymb}
\usepackage{xcolor}
\usepackage{multirow}
\usepackage{xspace}
\usepackage{hyperref}
\usepackage{booktabs}
\makeatletter
\DeclareRobustCommand\onedot{\futurelet\@let@token\@onedot}
\def\@onedot{\ifx\@let@token.\else.\null\fi\xspace}
\makeatother

\begin{document}
\sloppy

\title{ReGround-Surg: Reliability-Guided Anchor Grounding for Referring Surgical Video Segmentation}
\titlerunning{ReGround-Surg}

\author{Jiaxin Wen \and Ming Yin \and Lu Liu \and Zeyu Fu\thanks{Corresponding author: \url{z.fu@exeter.ac.uk}}}
\authorrunning{Wen et al.}
\institute{University of Exeter, Exeter, UK}

\maketitle

\begin{abstract}
Referring surgical video segmentation requires segmenting a target instrument or tissue region across video frames according to a natural language expression. Recent Segment Anything Model 2 (SAM2) based two-stage methods (e.g., ReSurgSAM2) first ground the referred target in an initial or selected frame, then propagate the selected mask via tracking. Although effective, their performance is highly sensitive to the quality of the initial grounded mask: once an incorrect anchor is selected, subsequent tracking tends to propagate the error. This issue is especially challenging in surgical videos due to visually similar instruments, occlusion, and complex tissue-tool interactions.
To address this issue, we propose \emph{ReGround-Surg}, a lightweight reliability-guided anchor
grounding framework to improve SAM2-based referring surgical video segmentation. It first predicts a text-conditioned spatial reliability map from the referring expression and current-frame visual features. The map is then reused in two complementary branches: a Gated Side Adapter enhances expression-relevant visual regions before text-to-vision fusion, while a Reliability-Weighted Vision-to-Text Attention module suppresses off-target visual evidence during prompt-token aggregation.
Experiments on Ref-EndoVis17 and Ref-EndoVis18 show consistent improvements over state-of-the-art methods across three evaluation splits with negligible speed reduction. Code is publicly available at \url{https://github.com/JiaxinWen1/ReGround-Surg}.

\keywords{Referring Surgical Instrument Segmentation \and SAM2 \and Reliability Gating}
\end{abstract}

\section{Introduction}
\label{sec:intro}

Surgical video segmentation plays a fundamental role in computer-assisted intervention by providing precise, pixel-level identification of surgical instruments and tissue regions. Existing methods focus on category-level segmentation, evolving from semantic segmentation~\cite{ronneberger2015unet} to instance-based approaches~\cite{gonzalez2020isinet}, transformer-based tracking~\cite{zhao2022trasetr}, and temporal consistency modeling~\cite{ayobi2023matis}. However, these methods are inherently limited in flexibility, as they cannot distinguish between instruments of the same type based on context or respond to free-form user queries.

To address this limitation, referring surgical video segmentation aims to segment surgical
instruments or tissue regions in a video according to a natural
language expression. Compared with category-level surgical instrument
segmentation, referring segmentation provides a more flexible interface because the target can be specified by object category, spatial location, action, or tissue relation. 
This is directly useful for interactive surgical video analysis, robotic assistance, and downstream skill assessment~\cite{maier2017surgical}. RSVIS~\cite{wang2024rsvis} establishes the foundations of this task by jointly learning video-level and instrument-level representations for cross-modal grounding. SurgRef~\cite{wei2026surgref} further shows that motion dynamics provide more robust grounding cues than static appearance alone, yet still relies on per-frame processing without exploiting video-level propagation.

\begin{figure}[t]
  \centering
  \includegraphics[width=0.65\linewidth]{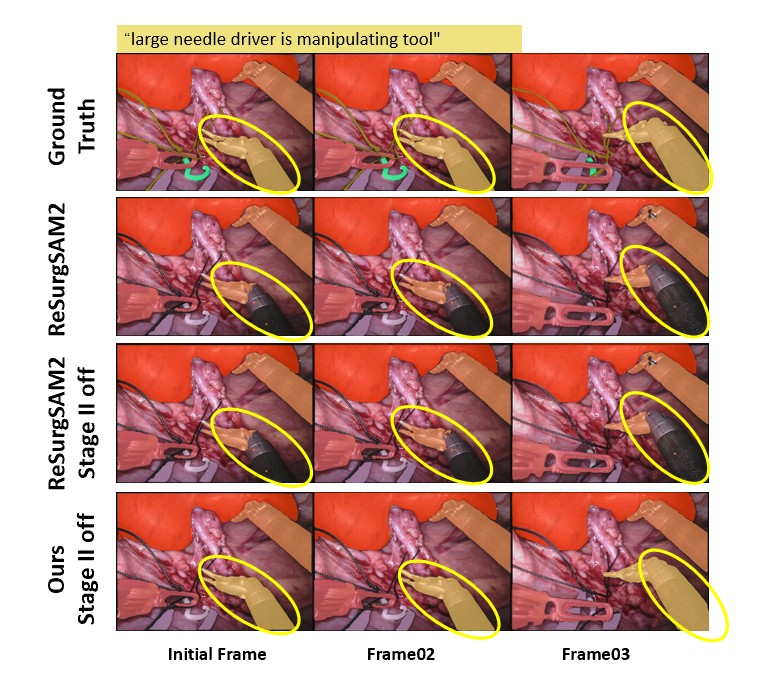}
  \caption{Visualisation of ReSurgSAM2 initial frame grounding and subsequent tracking failures 
under multi-instrument confusion, showing per-frame segmentation results 
for the expression \textit{``large needle driver is manipulating tool''} 
across four conditions: ground truth, ReSurgSAM2, ReSurgSAM2 (w/o Stage~\uppercase\expandafter{\romannumeral2}), and ReGround-Surg (Ours, w/o Stage~\uppercase\expandafter{\romannumeral2}).}
  \label{fig:motivation}
\end{figure}
Recent Segment Anything Model (SAM) based methods, such as ReSurgSAM2~\cite{liu2025resurgsam2}, have shown promising performance on referring surgical video segmentation. They commonly follow a two-stage detection-and-tracking paradigm: Stage~\uppercase\expandafter{\romannumeral1} grounds the referred target and selects a credible anchor frame with an anchor mask, while Stage~\uppercase\expandafter{\romannumeral2} propagates this mask to the remaining frames using the tracking memory of SAM2~\cite{ravi2024sam2}.
Despite this effective design, the two-stage pipeline remains vulnerable to errors in the initial grounding stage. As illustrated in Fig.~\ref{fig:motivation}, given the expression \textit{``large needle driver is manipulating tool''}, the baseline incorrectly grounds an adjacent instrument in the anchor frame, and the tracking stage subsequently propagates this error throughout the video. To determine whether this failure arises from tracking or from initial grounding, we evaluate a Stage~\uppercase\expandafter{\romannumeral2}-disabled variant that directly uses the Stage~\uppercase\expandafter{\romannumeral1} detection mask without temporal propagation. The incorrect prediction persists, as shown in the third row of Fig.~\ref{fig:motivation}, indicating that the error primarily stems from erroneous anchor selection in Stage~\uppercase\expandafter{\romannumeral1}.
This reveals a key limitation of current SAM2-based referring surgical video segmentation methods: strong tracking cannot compensate for inaccurate initial grounding. Once a wrong anchor mask is selected, the error is propagated throughout the video, especially under instrument similarity, occlusion, and tissue-tool interactions.

\begin{figure}[t]
  \centering
  \includegraphics[width=0.95\linewidth]{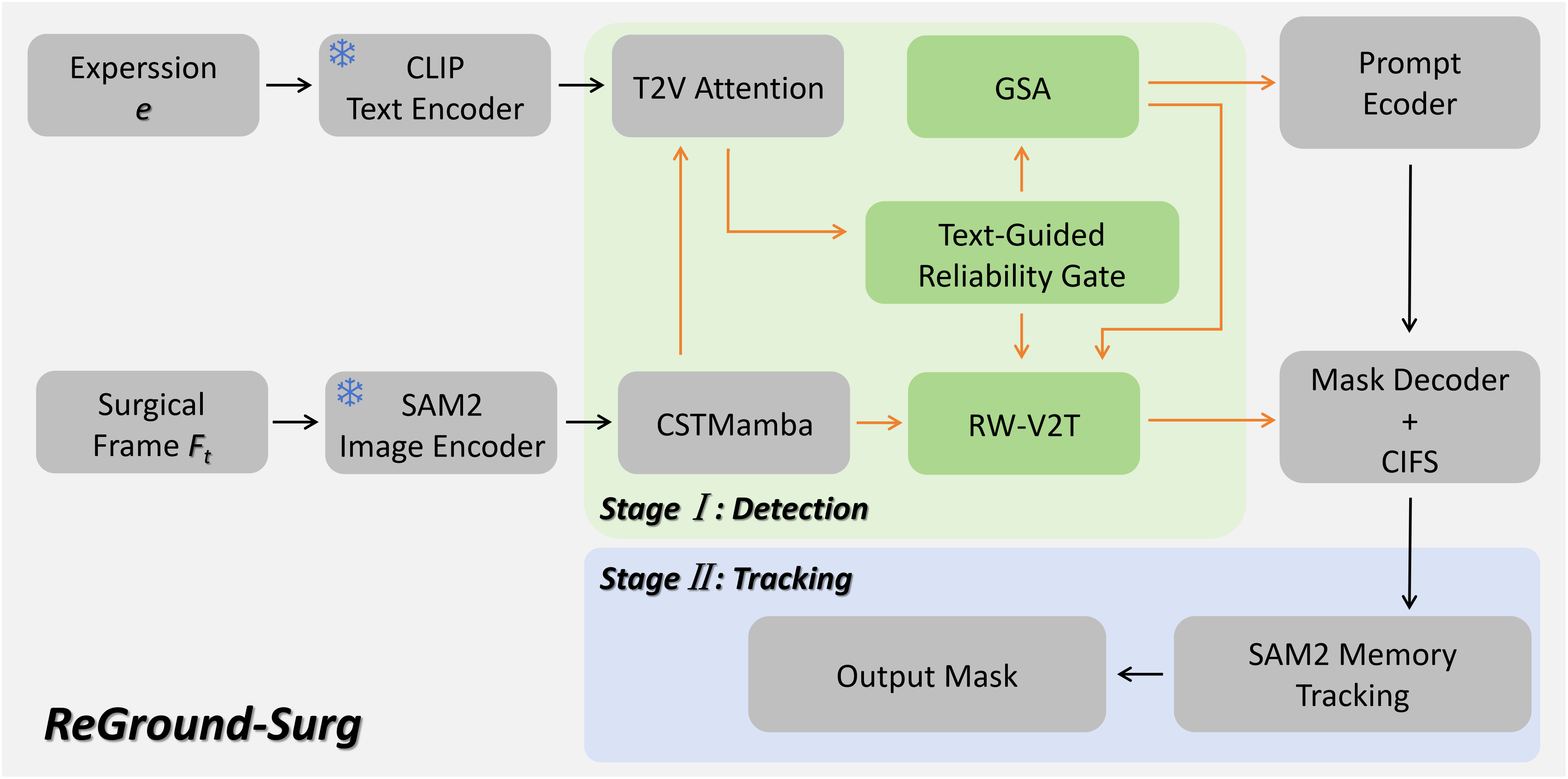}
  \caption{Overview of our proposed \textit{ReGround-Surg} for Referring Surgical Video Segmentation. }
  \label{fig:overview}
\end{figure}



These results highlight that robust referring surgical video segmentation demands reliable cross-modal grounding prior to tracking. Rather than modifying the SAM2 tracking stage, we focus on Stage~\uppercase\expandafter{\romannumeral1} and propose \emph{ReGround-Surg}, a lightweight reliability-guided anchor grounding framework that enhances expression-relevant regions and suppresses distractors during anchor selection, as illustrated in Fig.~\ref{fig:overview}.
Specifically, a Text-Guided Reliability Gate first predicts a text-conditioned spatial
reliability map from max-pooled noun token features and compressed current-frame visual features. This map is shared by two complementary modules: a Gated Side Adapter (GSA) that modulates visual features before text-to-vision attention, encouraging the model to focus on regions consistent with the referring expression; and a Reliability-Weighted Vision-to-Text (RW-V2T) attention mechanism that reuses the same reliability map as a multiplicative weight on visual keys when aggregating visual features into the prompt token, reducing the influence of off-target or unreliable regions on grounding. 

The main contributions of this paper are as follows: 
1) We analyse the initial grounding bottleneck in the current SAM2-based two-stage referring surgical video segmentation and provide quantitative and qualitative evidence that the inaccurate anchor mask selection is associated with persistent tracking failures in two-stage pipelines.
2) We propose \textit{ReGround-Surg}, in which a shared text-conditioned reliability map is estimated and used by two complementary modules: a GSA for visual feature modulation and an RW-V2T attention for suppressing off-target visual evidence during prompt-token aggregation.
3) Experiments on Ref-EndoVis17 and Ref-EndoVis18 show consistent improvements of $+3.77$, $+3.09$, and $+0.94$ over ReSurgSAM2 across three evaluation splits, with only $0.5\,\mathrm{M}$ additional parameters and negligible speed overhead.

\section{Related Work}
\label{sec:related}

\subsection{Foundation Models for Surgical Video Segmentation}
SAM~\cite{kirillov2023sam} and SAM2~\cite{ravi2024sam2} have demonstrated strong promptable and zero-shot segmentation capabilities, inspiring a range of surgical-specific adaptations. SurgicalSAM~\cite{yue2024surgicalsam} adapts SAM for surgical instrument segmentation via class-level prompts. SurgicalSAM2~\cite{liu2024surgicalsam2} further extends SAM2 to real-time surgical video segmentation via frame pruning strategies, while MA-SAM2~\cite{yin2025masamv2} introduces memory augmentation to improve long-sequence tracking stability. Despite these advances, these methods primarily rely on visual prompts such as points or boxes, limiting their ability to dynamically specify targets through natural language instructions in interactive surgical settings.

\subsection{Referring Surgical Video Segmentation}
Referring surgical video segmentation introduces natural language expressions to specify target instruments or tissue regions. 
In the general domain, referring video object segmentation (RVOS) methods such as ReferFormer~\cite{wu2022language}, MUTR~\cite{yan2024mutr}, and SAMWISE~\cite{cuttano2025samwise} have established strong baselines for language-guided video segmentation. However, these methods are designed for natural scenes and generalise poorly to surgical settings due to instrument similarity, occlusion, and challenging imaging conditions.
In the surgical domain, RSVIS~\cite{wang2024rsvis} introduces the referring surgical video instrument segmentation task and proposes VIS-Net with graph-based relation-aware representation learning. ReSurgSAM2~\cite{liu2025resurgsam2} combines CLIP text features~\cite{radford2021clip} with SAM2 video propagation in a two-stage detection-and-tracking framework, where an anchor mask is first grounded through credible initial frame selection and then propagated via SAM2 memory. More recently, SurgRef~\cite{wei2026surgref} argues that existing methods over-rely on static visual cues and predefined instrument names, and proposes motion-guided grounding to improve robustness under occlusion, semantic ambiguity, and cross-institutional scenarios.
Despite these advances, SAM2-based two-stage pipelines share a common assumption that the initial anchor grounding is reliable. When this assumption fails, the tracking stage propagates the error throughout the sequence with no mechanism for correction. This grounding bottleneck motivates the focus of our work.

\subsection{Cross-Modal Grounding and Feature Modulation}
To address the grounding bottleneck identified above, we draw on two complementary lines of work on cross-modal feature modulation.
LAVT~\cite{yang2022lavt} introduces text-guided spatial gating over intermediate visual feature maps to suppress irrelevant regions, but treats all spatial responses uniformly without distinguishing which are trustworthy under occlusion or instrument confusion. 
XMem~\cite{cheng2022xmem} demonstrates that explicitly assigning different weights to different elements is more effective than uniform aggregation, yet the reliability estimation relies solely on visual cues without any language conditioning to identify expression-relevant regions.
AdaptFormer~\cite{chen2022adaptformer} offers a complementary perspective, 
showing that zero-initialised side adapters can modulate pre-trained representations efficiently; yet it is designed for unimodal vision and lacks any mechanism for cross-modal alignment or reliability-aware suppression. 
These limitations suggest that neither line of work alone suffices for reliable cross-modal grounding under surgical complexity. Building on these insights, our work introduces lightweight reliability-anchor grounding to improve the effectiveness of SAM2-based referring surgical video segmentation.


\section{Method}
\label{sec:method}

\subsection{Preliminaries}
\label{sec:formulation}

Given a surgical video $\mathcal{V}=\{I_t\}_{t=1}^{T}$ and a
referring expression $\mathcal{Q}$, the goal is to predict binary
masks $\mathcal{M}=\{M_t\}_{t=1}^{T}$, where
$M_t \in \{0,1\}^{H \times W}$ indicates the referred instrument or
tissue region in frame $I_t$. 


Our method builds upon ReSurgSAM2~\cite{liu2025resurgsam2}, a two-stage framework for referring surgical video segmentation. In Stage~\uppercase\expandafter{\romannumeral1}, CLIP text features~\cite{radford2021clip} are fused with SAM2 visual features via a Cross-modal Spatial-Temporal Mamba (CSTMamba) module to produce per-frame segmentation masks. A Credible Initial Frame Selection (CIFS) strategy then identifies a reliable anchor frame $\hat{M}_a$ from a sliding window of high-confidence detections. In Stage~\uppercase\expandafter{\romannumeral2}, this anchor mask is propagated across the remaining frames via SAM2's memory-based tracking. Consequently, the accuracy of Stage~\uppercase\expandafter{\romannumeral1} grounding is critical, as it directly determines the quality of the entire output sequence.

The CSTMamba module fuses text and visual features through a \texttt{TwoWayTokenTransformer} with $L{=}3$ stacked blocks, each performing 
text self-attention, a Mamba temporal layer~\cite{gu2023mamba}, text-to-vision (T2V) cross-attention, an MLP step, and vision-to-text (V2T) cross-attention. We identify two structural limitations that undermine grounding reliability. In the \emph{T$\to$V direction}, all $H{\times}W$ spatial positions are treated uniformly, allowing diffuse activations that the mask decoder cannot resolve. In the \emph{V$\to$T direction}, all visual tokens contribute equally to the prompt token \texttt{[CLS']}, regardless 
of occlusion or background clutter, potentially corrupting the grounding signal. As confirmed by Fig.~\ref{fig:motivation}, disabling Stage~\uppercase\expandafter{\romannumeral2} does not resolve grounding failures, directly implicating CSTMamba as the source of error. 
To address these gaps, we propose a lightweight reliability-guided framework for improving Stage 1 grounding in SAM2-based referring surgical video segmentation as illustrated in Fig.~\ref{fig:overview}.

\begin{figure}[t]
  \centering
  \includegraphics[width=0.9\linewidth]{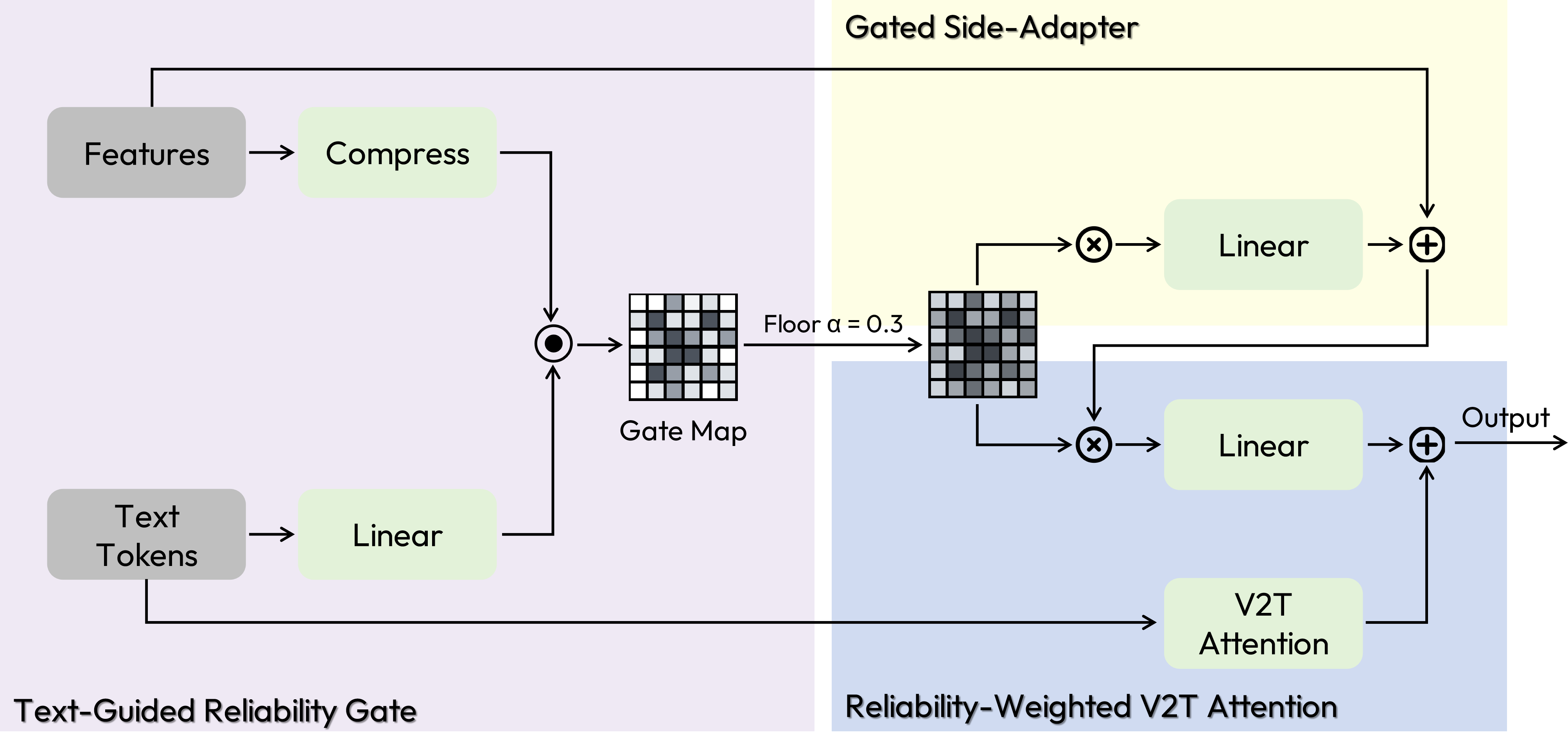}
  \caption{Architecture of the proposed Text-Guided Reliability Gate, 
Gated Side Adapter (GSA), and Reliability-Weighted V2T Attention (RW-V2T).}
  \label{fig:gate}
\end{figure}

\subsection{Reliability-Guided Anchor
Grounding}


\noindent\textbf{Text-Guided Reliability Gate.}\label{sec:gate}
The Text-Guided Reliability Gate is the shared core of our method, as illustrated in Fig.~\ref{fig:gate}. It predicts a spatial reliability map $G \in [0,1]^{H \times W}$ that indicates, for each spatial position, the degree of alignment between the visual content and the referring expression. This map is then reused by both the Gated Side Adapter and the Reliability-Weighted V2T Attention.

Let $\mathbf{Q} \in \mathbb{R}^{N \times C}$ denote the text token sequence output from the self-attention layer, where $N$ is the number of text tokens and $C$ is the feature dimension. To capture the most discriminative text features (which typically correspond to instrument-related nouns), inspired by LAVT~\cite{yang2022lavt}, we apply element-wise max-pooling over all text tokens:

\begin{equation}
  \mathbf{n}_c = \max_{i=1}^{N}\,\mathbf{Q}_{:,i,c}
  \;\in\; \mathbb{R}^{B \times C},
  \label{eq:noun}
\end{equation}

Both $\mathbf{n}$ and the current-frame visual features
$\mathbf{K}_\mathrm{cur} \in \mathbb{R}^{B \times C \times H \times W}$
are projected to a 64-dimensional bottleneck:
$\hat{\mathbf{n}} = W_\mathrm{gate}\mathbf{n}$ and
$\hat{\mathbf{K}} = \mathrm{Conv}_{1\times1}(\mathbf{K}_\mathrm{cur})$.
The reliability map is then:
\begin{equation}
  G = \sigma\!\left(
    \textstyle\sum_{d=1}^{64}
    \hat{\mathbf{K}}_{:,d,:,:}\cdot\hat{\mathbf{n}}_{:,d}
  \right)
  \;\in\;[0,1]^{B\times 1\times H\times W},
  \label{eq:gate}
\end{equation}
where $G{\approx}1$ indicates strong text--vision alignment and
$G{\approx}0$ indicates poor alignment or distractor regions.
$G$ is computed independently in each of the three
\texttt{TwoWayTokenAttentionBlock} layers with separate parameters,
allowing it to adapt to progressively refined representations.

\noindent\textbf{Gated Side Adapter (GSA).} \label{sec:mod1}
The reliability map is used to modulate visual features before T2V
cross-attention via a soft residual gate with floor $\alpha \in [0,1)$, where $\alpha{=}0.3$ is selected by grid search (Table~\ref{tab:alpha}):
\begin{equation}
  \tilde{\mathbf{K}} = \hat{\mathbf{K}}\cdot\bigl(\alpha+(1-\alpha)G\bigr).
  \label{eq:modulate}
\end{equation}
We use residual modulation rather than direct masking to avoid suppressing potentially useful contextual information, especially when the expression is ambiguous, or the target is partially
occluded~\cite{oktay2018attention}.
The modulated features are expanded back to $C$ channels and added as
a zero-initialised~\cite{chen2022adaptformer} residual:
\begin{equation}
  \mathbf{K}_\mathrm{cur}' =
    \mathrm{LayerNorm}\!\bigl(
      \mathbf{K}_\mathrm{cur}
      + \mathrm{Conv}_{1\times1}^{\uparrow}(\tilde{\mathbf{K}})
    \bigr).
  \label{eq:residual}
\end{equation}
Zero-initialisation of $\mathrm{Conv}_{1\times1}^{\uparrow}$ ensures the adapter is functionally identical to the baseline at step zero, preserving pre-trained SAM2 representations. Only current-frame features are modulated; reference-frame features selected by CIFS are left unchanged, since gating historical memory would risk undermining the stable temporal anchor.


\noindent\textbf{Reliability-Weighted V2T Attention (RW-V2T).}\label{sec:mod2}
Standard V2T cross-attention lets all visual tokens contribute equally
to the prompt token, regardless of occlusion or background.
We incorporate the reliability map $G$ as a multiplicative weight
on the visual keys, so that expression-relevant positions contribute
more strongly to prompt-token aggregation.
The reliability-weighted keys are computed as:
\begin{equation}
  \mathbf{K}_\mathrm{v2t} =
    \mathrm{LayerNorm}\!\bigl(
      \mathbf{K} + W_v(\mathbf{K} \cdot v_\mathrm{full})
    \bigr),
  \label{eq:vis_proj}
\end{equation}
where $v_\mathrm{full} = [\,(\alpha + (1-\alpha)G_\mathrm{flat})\;\|\;\mathbf{1}_{(f-1)HW}\,]$, $f$ denotes the total number of frames (one current frame plus $f{-}1$ reference frames),
and $W_v \in \mathbb{R}^{C \times C}$ is a learnable linear projection
initialised as the identity matrix.
The residual is added to the original $\mathbf{K}$ rather than to the
weighted version, so downstream visual features retain their full
unattenuated content (information-lossless design).
No explicit supervision is imposed on the reliability gate; it is
learned implicitly through the segmentation loss.

\noindent\textbf{Training and Inference.}
\label{sec:training}
To preserve the pre-trained SAM2 representation, the SAM2 image encoder and tracking memory module are frozen throughout training.
Only the \texttt{CrossModalFusionModule} (original CSTMamba weights 
plus the $0.5\,\mathrm{M}$ parameters of the two proposed modules)
is updated. 
The model is trained with the same composite loss as ReSurgSAM2~\cite{liu2025resurgsam2}:
\begin{equation}
  \mathcal{L} = 20\mathcal{L}_\mathrm{mask}
              + \mathcal{L}_\mathrm{dice}
              + \mathcal{L}_\mathrm{IoU}
              + \mathcal{L}_\mathrm{cls},
  \label{eq:loss}
\end{equation}
where $\mathcal{L}_\mathrm{mask}$ is binary cross-entropy,
$\mathcal{L}_\mathrm{dice}$ provides a region-overlap objective,
$\mathcal{L}_\mathrm{IoU}$ directly optimises the confidence score
consumed by CIFS, and $\mathcal{L}_\mathrm{cls}$ supervises the
occlusion prediction head.

For inference, the proposed modules are used only in Stage~\uppercase\expandafter{\romannumeral1} detection. The resulting anchor mask is then passed to the original SAM2 tracking stage unchanged, making our method fully plug-and-play with respect to Stage~\uppercase\expandafter{\romannumeral2}.

\section{Experiments}
\label{sec:experiments}

\subsection{Datasets and Experimental Details}
\label{sec:datasets}

\noindent\textbf{Datasets.}
We evaluate on Ref-EndoVis17 and Ref-EndoVis18~\cite{wang2024rsvis},
following the same splits as ReSurgSAM2~\cite{liu2025resurgsam2}.
Both datasets build on EndoVis17~\cite{allan2019endovis17} and
EndoVis18~\cite{allan2020endovis18}, reannotated with instance-specific
labels. Ref-EndoVis18 includes tissue-specific annotations.
Table~\ref{tab:datasets} summarises the statistics.
\begin{table}[t]
\centering
\small
\setlength{\tabcolsep}{4pt}
\caption{Dataset statistics.}
\label{tab:datasets}
\begin{tabular}{lcccccccc}
\toprule
\multirow{2}{*}{Dataset}
  & \multicolumn{4}{c}{Training}
  & \multicolumn{4}{c}{Testing} \\
\cmidrule(lr){2-5}\cmidrule(lr){6-9}
  & Seq. & Frame & Object & Pair
  & Seq. & Frame & Object & Pair \\
\midrule
Ref-EndoVis17 (tool)   & 7 & 2100 & 20 & 4873 & 3 & 900  & 10 & 2265 \\
Ref-EndoVis18 (tool)   & 11 & 1639 & 34 & 3787 & 4 & 596 & 15 & 1384 \\
Ref-EndoVis18 (tissue) & 11 & 1639 & 25 & 2995 & 4 & 596 &  7 &  807 \\
\bottomrule
\end{tabular}
\end{table}

\noindent\textbf{Metrics.}
We adopt region similarity $\mathcal{J} = \frac{|M \cap G|}{|M \cup G|}$, contour accuracy $\mathcal{F} = \frac{2P_c R_c}{P_c + R_c}$~\cite{perazzi2016benchmark}, and their mean $\mathcal{J\&F} = \frac{1}{2}(\mathcal{J} + \mathcal{F})$ following the RVOS evaluation protocol~\cite{wu2022language}, where $M$ and $G$ denote the predicted mask and ground-truth mask respectively, and $P_c$, $R_c$ are the contour precision and recall. All metrics are higher-is-better.
To better understand the effect of Stage~\uppercase\expandafter{\romannumeral1} detection quality, we additionally report Initial Grounding metrics: $\mathcal{J}$, $\mathcal{F}$, and $\mathcal{J\&F}$ computed on the CIFS-selected anchor frame before SAM2 tracking. We further introduce AnchorAcc, defined as the fraction of anchor masks whose IoU with ground truth exceeds $0.5$:
\begin{equation}
  \mathrm{AnchorAcc} =
  \frac{1}{S}\sum_{i=1}^{S}
  \mathbb{1}
  \Bigl[
    \mathrm{IoU}
    \bigl(
      \hat{M}_a^{(i)},
      M_a^{(i)}
    \bigr)
    > 0.5
  \Bigr].
  \label{eq:anchoracc}
\end{equation}
where $S$ is the number of video-text pairs, $\hat{M}_a^{(i)}$ is the predicted anchor frame mask, and $M_a^{(i)}$ is the corresponding ground-truth mask.

\noindent\textbf{Implementation Details.}
\label{sec:impl}
We use the Hiera-Small backbone~\cite{ryali2023hiera} initialised from the official ReSurgSAM2 checkpoint~\cite{liu2025resurgsam2}. Input resolution is $512{\times}512$. The SAM2 image encoder, CLIP text encoder~\cite{radford2021clip}, SAM2 mask decoder, and all Stage~\uppercase\expandafter{\romannumeral2} components are frozen; only the \texttt{CrossModalFusionModule} is updated. The model is trained for 60 epochs with a batch size of 16 (8~GPUs, 2~per~GPU) and the AdamW optimiser. The original CSTMamba parameters use a learning rate of $5{\times}10^{-5}$, while our newly introduced adapter parameters use a larger rate of $3{\times}10^{-4}$ to compensate for their zero/identity initialisation~\cite{chen2022adaptformer}. The schedule follows linear warm-up (30\%) then cosine decay; gradients are clipped to $\ell_2$ norm $0.1$. For inference, we adopt the default ReSurgSAM2 hyperparameters: $\delta_o{=}0.9$, $\delta_{iou}{=}0.7$, $\gamma_{iou}{=}0.95$, $N_w{=}5$, $N_p{=}5$, $N_l{=}4$ (see~\cite{liu2025resurgsam2} for definitions).

\subsection{Main Results}
\label{sec:results}

\begin{table}[t]
\centering
\small
\setlength{\tabcolsep}{1.5pt}
\caption{Comparison with state of the art on Ref-EndoVis17 and
  Ref-EndoVis18. $\dagger$: offline method.
  \textbf{Bold}: best; \underline{underline}: second best.
  $\Delta$: gain over ReSurgSAM2. FPS measured on NVIDIA A100-SXM4-80GB.}
\label{tab:main}
\begin{tabular}{lccccccccc|c}
\toprule
& \multicolumn{3}{c}{EV17 (tool)}
& \multicolumn{3}{c}{EV18 (tool)}
& \multicolumn{3}{c}{EV18 (tissue)} & \\
\cmidrule(lr){2-4}\cmidrule(lr){5-7}\cmidrule(lr){8-10}
Method
  & $\mathcal{J\&F}$ & $\mathcal{J}$ & $\mathcal{F}$
  & $\mathcal{J\&F}$ & $\mathcal{J}$ & $\mathcal{F}$
  & $\mathcal{J\&F}$ & $\mathcal{J}$ & $\mathcal{F}$ & FPS \\
\midrule
ReferFormer$^\dagger$~\cite{wu2022language}
  & 52.64 & 53.06 & 52.22 & 65.43 & 66.14 & 64.72
  & 66.37 & 72.18 & 60.56 & --- \\
MUTR$^\dagger$~\cite{yan2024mutr}
  & 54.09 & 53.80 & 54.38 & 66.92 & 67.71 & 66.13
  & 66.14 & 72.37 & 59.91 & --- \\
RSVIS~\cite{wang2024rsvis}
  & 61.76 & 62.06 & 61.46 & 71.28 & 71.85 & 70.71
  & 70.64 & 76.26 & 65.02 & --- \\
OnlineRefer~\cite{wu2023onlinerefer}
  & 60.35 & 60.92 & 59.78 & 70.17 & 70.93 & 69.41
  & 68.83 & 74.51 & 63.15 & --- \\
RefSAM~\cite{li2024refsam}
  & 63.56 & 63.84 & 63.28 & 72.86 & 73.40 & 72.32
  & 71.90 & 77.48 & 66.32 & --- \\
\midrule
ReSurgSAM2~\cite{liu2025resurgsam2}
  & \underline{77.73} & \underline{77.77} & \underline{77.69}
  & \underline{80.62} & \underline{80.94} & \underline{80.31}
  & \underline{75.09} & \underline{80.93} & \underline{69.25}
  & \underline{54.2} \\
  \rowcolor{gray!10}
Ours
  & \textbf{81.50} & \textbf{81.48} & \textbf{81.52}
  & \textbf{83.71} & \textbf{83.83} & \textbf{83.60}
  & \textbf{76.03} & \textbf{82.90} & \textbf{69.33}
  & \textbf{53.9} \\
\midrule
$\Delta$
  & \textcolor{teal}{+3.77} & \textcolor{teal}{+3.71} & \textcolor{teal}{+3.83}
  & \textcolor{teal}{+3.09} & \textcolor{teal}{+2.89} & \textcolor{teal}{+3.29}
  & \textcolor{teal}{+0.94} & \textcolor{teal}{+1.97} & \textcolor{teal}{+0.08}
  & \textcolor{teal}{$-$0.3} \\
\bottomrule
\end{tabular}
\end{table}
\begin{table}[t]
\centering
\small
\setlength{\tabcolsep}{6pt}
\caption{Module ablation on Ref-EndoVis18 (tool).}
\label{tab:component}
\begin{tabular}{cc|ccc}
\toprule
GSA & RW-V2T & $\mathcal{J\&F}$ & $\mathcal{J}$ & $\mathcal{F}$ \\
\midrule
           &            & 80.62 & 80.94 & 80.31 \\
\checkmark &            & 82.43 & 82.68 & 82.18 \\
           & \checkmark & 81.22 & 81.40 & 81.04 \\
\checkmark & \checkmark & \textbf{83.71} & \textbf{83.83} & \textbf{83.60} \\
\bottomrule
\end{tabular}
\end{table}

Table~\ref{tab:main} compares our method against prior work on all three evaluation splits. Our method achieves the best performance on every metric across all datasets. Compared to the strong ReSurgSAM2 baseline, our method further improves $\mathcal{J\&F}$ by $+3.77$ and $+3.09$ on the two tool splits, demonstrating clear benefits from reliability-aware grounding. The gain on the tissue split is more modest ($+0.94$), which we attribute to the spatially diffuse nature of tissue regions: their boundaries are less well-defined than rigid instruments, making the reliability gate's spatial focusing less impactful. Nevertheless, our method still establishes a new state of the art on this challenging split. We do not include SurgRef~\cite{wei2026surgref} in the quantitative comparison, as its pretrained checkpoints have not been publicly released at the time of submission.

\subsection{Ablation Studies}
\label{sec:ablation}

All ablation experiments are conducted on Ref-EndoVis18 (tool) to examine the contribution of each design choice.

\noindent\textbf{Module contribution.}
Table~\ref{tab:component} reports the contribution of each proposed module. 
Since both GSA and RW-V2T consume the reliability map $G$ produced by the 
Text-Guided Reliability Gate, the Gate is kept enabled throughout; its 
internal design is dissected separately in Table~\ref{tab:gate_design}.
Individually, GSA improves $\mathcal{J\&F}$ by $+1.81$ and RW-V2T by 
$+0.60$, confirming that injecting reliability awareness into either attention direction is beneficial on its own. More importantly, combining the two modules yields a $+3.09$ gain, which exceeds the sum of their 
individual contributions by $0.68$ points. This indicates that GSA and RW-V2T are complementary to each other. By anchoring both modules to the same reliability map, our design enables coordinated reliability-aware modulation in both 
attention directions.

\begin{table}[t]
\centering

\begin{minipage}[t]{0.48\linewidth}
\centering
\small
\setlength{\tabcolsep}{5pt}

\caption{Gate design ablation on Ref-EndoVis18 (tool).}
\label{tab:gate_design}
\begin{tabular}{lc}
\toprule
Gate design & $\mathcal{J\&F}$ \\
\midrule
No gate (baseline)               & 80.62 \\
Static learned gate              & 80.79 \\
Visual-only gate                 & 81.02 \\
Text-guided gate (no residual)   & 81.94\\
Text-guided residual gate (ours) & \textbf{82.43} \\
\bottomrule
\end{tabular}
\end{minipage}
\hfill
\begin{minipage}[t]{0.48\linewidth}
\centering
\small
\setlength{\tabcolsep}{5pt}
\caption{V2T attention design ablation on Ref-EndoVis18 (tool).}
\label{tab:v2t_design}
\begin{tabular}{lc}
\toprule
V2T design & $\mathcal{J\&F}$ \\
\midrule
Vanilla V2T (baseline)       & 80.62 \\
$V \odot G$ + mixing         & 80.95 \\
$V \odot G$ + mixing + residual (ours) & \textbf{81.22} \\
\bottomrule
\end{tabular}
\end{minipage}

\end{table}
\begin{table}[t]
\centering
\begin{minipage}[t]{0.45\linewidth}
\centering
\small
\setlength{\tabcolsep}{1.7pt}
\caption{Effect of gate floor $\alpha$ on Ref-EndoVis18 (tool).}
\label{tab:alpha}
\begin{tabular}{l|ccccc}
\toprule
$\alpha$ & 0.0 & 0.1 & 0.3 & 0.5 & 1.0 \\
\midrule
$\mathcal{J\&F}$ & 81.94 & 82.85 & \textbf{83.71} & 83.12 & 81.35 \\
\bottomrule
\end{tabular}
\end{minipage}
\hfill
\begin{minipage}[t]{0.5\linewidth}
\centering
\small
\setlength{\tabcolsep}{5pt}
\caption{Effect of initialisation strategy on Ref-EndoVis18 (tool).}
\label{tab:init}
\begin{tabular}{l|cc}
\toprule
Initialisation & $\mathcal{J\&F}$  \\
\midrule
Random init               & 79.53            \\
Zero/identity init (ours) & \textbf{83.71}  \\
\bottomrule
\end{tabular}
\end{minipage}
\end{table}

\noindent\textbf{Gate design.}
Table~\ref{tab:gate_design} dissects the reliability gate design. 
A static learned gate or visual-only gate yields only marginal gains 
($+0.17$ and $+0.40$ over the no-gate baseline), confirming that text 
conditioning is the key contributor. Removing the residual structure 
from our text-guided gate further drops $\mathcal{J\&F}$ from $82.43$ 
to $81.94$, showing that soft residual modulation outperforms direct gating.

\noindent\textbf{V2T attention design.}
Table~\ref{tab:v2t_design} examines the RW-V2T design.
Value scaling with channel mixing yields a modest gain ($+0.33$),
and adding the identity-initialised residual projection brings a
further improvement ($+0.27$). The residual connection stabilises
training by preserving original visual content.

\noindent\textbf{Gate floor $\alpha$.}
Table~\ref{tab:alpha} reports sensitivity to the soft floor parameter 
$\alpha$ in Eq.~\ref{eq:modulate}. Performance follows an inverted-U pattern: 
$\alpha{=}0$ aggressively suppresses low-confidence regions and removes 
useful contextual cues ($81.94$), whereas $\alpha{=}1$ degenerates into 
uniform weighting and disables reliability gating altogether ($81.35$). 
$\alpha{=}0.3$ achieves the best trade-off between distractor suppression 
and context preservation ($83.71$).

\noindent\textbf{Initialisation strategy.}
Table~\ref{tab:init} compares random initialisation against our zero/identity 
initialisation for the newly introduced adapter parameters. Random 
initialisation severely degrades performance ($-4.18$ $\mathcal{J\&F}$), 
confirming that preserving the pre-trained CSTMamba representation at 
initialisation is critical for stable fine-tuning of the lightweight 
adapter modules~\cite{chen2022adaptformer}.

\subsection{Further Analysis}
\label{sec:grounding}
\noindent\textbf{Initial Grounding Analysis.}
To verify that our improvements stem from better Stage~\uppercase\expandafter{\romannumeral1} grounding rather than from incidental interactions with the tracking stage, we report 
grounding quality on the CIFS-selected anchor frame before SAM2 tracking. 
Table~\ref{tab:grounding} reports grounding quality on the CIFS-selected 
anchor frame before SAM2 tracking. Our method improves Init.\ $\mathcal{J\&F}$ 
by $+9.52$ and AnchorAcc by $+8.98$, which translate into a $+3.09$ gain 
in final video $\mathcal{J\&F}$. The smaller final gain reflects the 
smoothing effect of SAM2 tracking, which compensates for minor noise, but 
cannot recover from anchor errors, reinforcing the importance of reliable 
Stage~\uppercase\expandafter{\romannumeral1} grounding.
\begin{table}[t]
\centering
\small
\setlength{\tabcolsep}{5pt}
\caption{Initial grounding analysis on Ref-EndoVis18 (tool).
  Metrics are computed on the CIFS-selected anchor frame before
  SAM2 tracking.}
\label{tab:grounding}
\begin{tabular}{l|ccccc}
\toprule
Method
  & Init.\ $\mathcal{J}$
  & Init.\ $\mathcal{F}$
  & Init.\ $\mathcal{J\&F}$
  & AnchorAcc 
  & Final $\mathcal{J\&F}$ \\
\midrule
ReSurgSAM2~\cite{liu2025resurgsam2}
  & 63.30 & 62.49 & 62.90 & 76.19 & 80.62 \\
  \rowcolor{gray!10}
ReGround-Surg (Ours)
  & 72.31 & 72.52 & 72.42 & 85.17 & \textbf{83.71} \\
\midrule
$\Delta$
  & \textcolor{teal}{+9.01}
  & \textcolor{teal}{+10.03}
  & \textcolor{teal}{+9.52}
  & \textcolor{teal}{+8.98}
  & \textcolor{teal}{+3.09} \\
\bottomrule
\end{tabular}
\end{table}
\begin{figure}[t]
  \centering
  \includegraphics[width=0.6\linewidth]{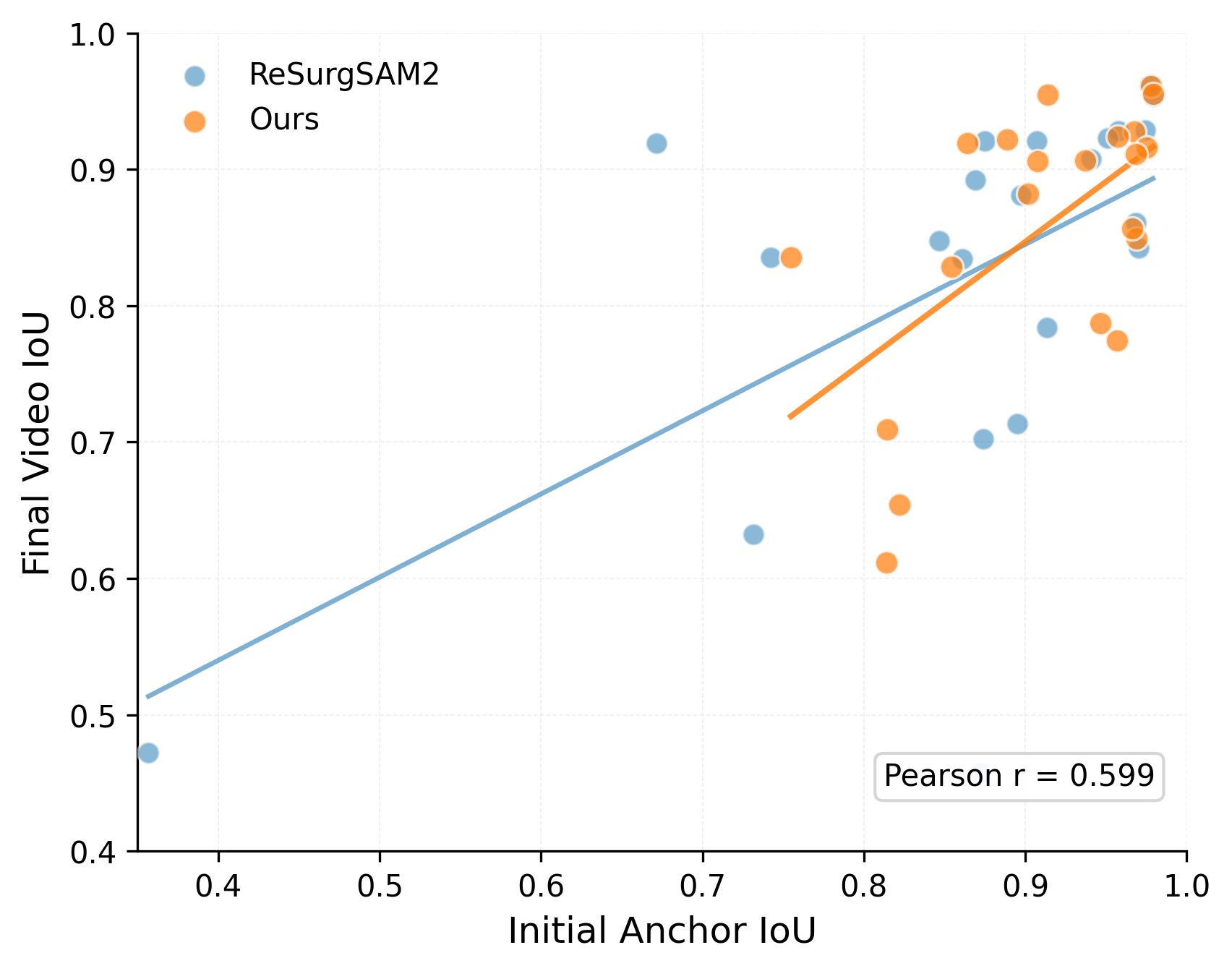}
  \caption{Initial anchor IoU vs.\ final video IoU on Ref-EndoVis17 and
  Ref-EndoVis18 (tool). Each point corresponds to one
  video-text pair. }
  \label{fig:scatter}
\end{figure}
Fig.~\ref{fig:scatter} plots the initial anchor IoU against the final video 
$\mathcal{J\&F}$ on Ref-EndoVis17 and Ref-EndoVis18 (tool). The clear 
positive trend confirms that a correct anchor is necessary, though not 
sufficient for high final performance, as the remaining variance arises 
from occlusion and scene complexity during tracking.

\noindent\textbf{Difficult-Case Analysis.}
\label{sec:difficult}
Table~\ref{tab:difficult} stratifies the Ref-EndoVis18 test
cases by scene complexity. Consistent improvements are observed across
all challenging scenarios. Gains are particularly pronounced for
\emph{same-type instrument confusion} ($+5.62$),
followed by \emph{partial occlusion} ($+4.17$) and
\emph{multi-instrument scenes} ($+4.15$). These results suggest that the proposed reliability gate is especially
beneficial when visually similar distractors compete with the referred
target, improving robustness to ambiguous surgical conditions.
\begin{table}[t]
\centering
\small
\setlength{\tabcolsep}{5pt}
\caption{Difficult-case analysis on Ref-EndoVis18,
  stratified by scene type ($\mathcal{J\&F}$).}
\label{tab:difficult}
\begin{tabular}{l|ccc}
\toprule
Case type & ReSurgSAM2 & Ours & Gain \\
\midrule
Multiple instruments    & 75.33 & \textbf{79.48} & \textcolor{teal}{+4.15} \\
Same-type instruments   & 71.39 & \textbf{77.01 }& \textcolor{teal}{+5.62} \\
Partial occlusion       & 75.43 & \textbf{79.60 }& \textcolor{teal}{+4.17} \\
\bottomrule
\end{tabular}
\end{table}
\begin{figure}[t]
  \centering
  \includegraphics[width=\linewidth]{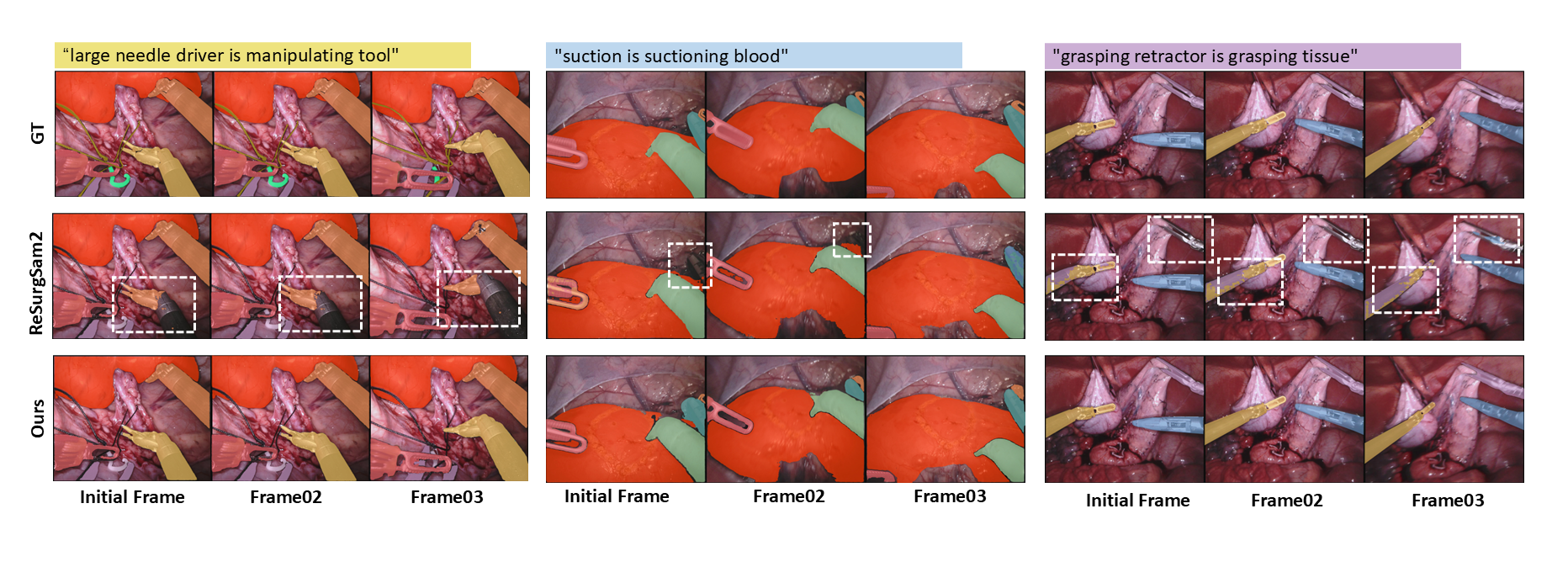}
  \caption{Qualitative comparison on
    Ref-EndoVis18. Rows (top to bottom): ground truth;  ReSurgSAM2 prediction; ours.
    Columns: multi-instrument confusion; partial occlusion; complete
    grounding failure.}
  \label{fig:qualitative}
\end{figure}

\noindent\textbf{Qualitative Analysis.}
\label{sec:qualitative}
Fig.~\ref{fig:qualitative} presents qualitative comparisons on Ref-EndoVis18 across three representative failure
modes: spatial confusion under multi-instrument scenes, first-frame
failure under partial occlusion, and a complete grounding failure case.
In the multi-instrument case, the baseline drifts onto an adjacent
instrument, while the reliability gate produces higher responses around
the referred region and lower responses around competing distractors.
In the occlusion case, RW-V2T appears to reduce the influence of
occluded or unreliable regions during prompt-token aggregation, enabling
better first-frame localisation. 
\section{Conclusion}
\label{sec:conclusion}

We presented ReGround-Surg, a lightweight reliability-guided anchor
grounding framework to enhance the referring surgical video segmentation. Motivated by the observation
that SAM2-based two-stage pipelines are highly sensitive to initial
grounding errors, our method improves Stage~\uppercase\expandafter{\romannumeral1} detection through a
Gated Side Adapter and Reliability-Weighted V2T attention.
Both modules share a single text-conditioned reliability map and are
inert at initialisation, and require no changes to the SAM2 encoder,
decoder, or tracking components.
Experiments on Ref-EndoVis17 and Ref-EndoVis18 show consistent
$\mathcal{J\&F}$ improvements of $+3.77$, $+3.09$, and $+0.94$ over
ReSurgSAM2 with only $0.5\,\mathrm{M}$ additional parameters.
Initial grounding analysis, gate visualisation, and difficult-case
breakdown further support the importance of addressing grounding
failure at its source, particularly under multi-instrument
confusion and partial occlusion.

Nevertheless, the current method has limitations: the reliability gate may produce weaker improvements on spatially diffuse targets such as tissue regions, and the gate floor $\alpha$ requires manual tuning across different surgical domain. Moreover, since Stage~\uppercase\expandafter{\romannumeral2} tracking remains unchanged, residual errors arising from memory drift or long-term occlusion after the anchor frame are not addressed by our current design. Future work will explore learnable gate parameters and extending reliability-aware modulation into Stage~\uppercase\expandafter{\romannumeral2} memory selection to further reduce propagation errors. 
\bibliographystyle{splncs04}
\bibliography{main}

\end{document}